\documentclass[11pt]{article}

\usepackage[final]{acl}

\usepackage{times}
\usepackage{latexsym}

\usepackage[T1]{fontenc}
\usepackage[utf8]{inputenc}

\usepackage{microtype}

\usepackage{inconsolata}

\usepackage{graphicx}
\usepackage{subcaption}
\usepackage{booktabs} 
\usepackage{amsmath} 
\usepackage{float}
\usepackage{fvextra}

\title{Tracing Recurrent and Emergent Hallucination Patterns in Decomposed Text-to-SQL Generation}

\author{
  \textbf{Hayden Moore} \\
  The Pennsylvania State University \\
  \texttt{hmm5731@psu.edu}
  \And
  \textbf{Abhinav Verma} \\
  The Pennsylvania State University \\
  \texttt{verma@psu.edu}
  \AND
  \textbf{Anshuman Dutt} \\
  Microsoft \\
  \texttt{anshuman.dutt@microsoft.com}
}
\begin{document}
\maketitle
\begin{abstract}
Large language models can achieve strong Text-to-SQL benchmark performance while still producing schema and query-logic errors. Most evaluations characterize such failures only in the final generated query, making it difficult to determine whether a related error was also observed in an earlier, simpler step or appears only at the final stage. We present a drill-down evaluation framework that decomposes Text-to-SQL queries into progressively more complex executable sub-queries and corresponding natural-language questions, allowing rule-based hallucination categories to be examined across a generation path. We distinguish \textit{recurrent hallucinations}, in which the same rule-based category is observed in an earlier step and again in the final query, from \textit{emergent hallucinations}, in which the category is observed at the final step without an earlier detection in the same path. Across five LLMs evaluated on an expanded BIRD-mini benchmark, several categories show appreciable recurrence probabilities, while clause abuse shows comparatively higher conditional emergence. Under the conversational-history condition, recurrence probabilities are generally higher and emergence probabilities are generally lower. These observations indicate that the two conditional measures do not move uniformly in the same direction under additional context and motivate trajectory-level analysis in addition to final-output evaluation. Code and evaluation artifacts are available in the project repository.\footnote{\url{https://github.com/HaydenMM/bird-mini-drilldown}}
\end{abstract}

\section{Introduction}

Text-to-SQL systems allow users to query structured databases through natural language rather than directly writing SQL. LLM-based approaches are increasingly used for this task because they can map natural-language requests to executable programs \citep{li2023bird, Chen2024CodeSTBB, Hong2024NextGenerationDIE}. Even when benchmark performance is relatively strong, generated SQL can still contain schema, value, join, clause, or other logic errors \citep{huang2023survey_hallucination, qu2024before_generation_align_it, Zhang2024FinSQLMLB}. In this paper, we operationalize a subset of these errors as hallucination categories using explicit rule-based comparisons against the database schema and reference SQL. Such errors can matter functionally because a syntactically valid query may execute while returning a result different from the reference query.

Most Text-to-SQL evaluation focuses on the final generated query. This provides a useful measure of task success but does not describe whether the same rule-detected error category was also present in earlier decomposed steps. A final-step hallucination may be preceded by an earlier instance of the same category, or it may be observed only at the final stage. Distinguishing these cases provides an additional descriptive view of hallucination behavior across a structured generation trajectory.

\begin{figure*}[t]
\centering
\includegraphics[width=0.96\textwidth]{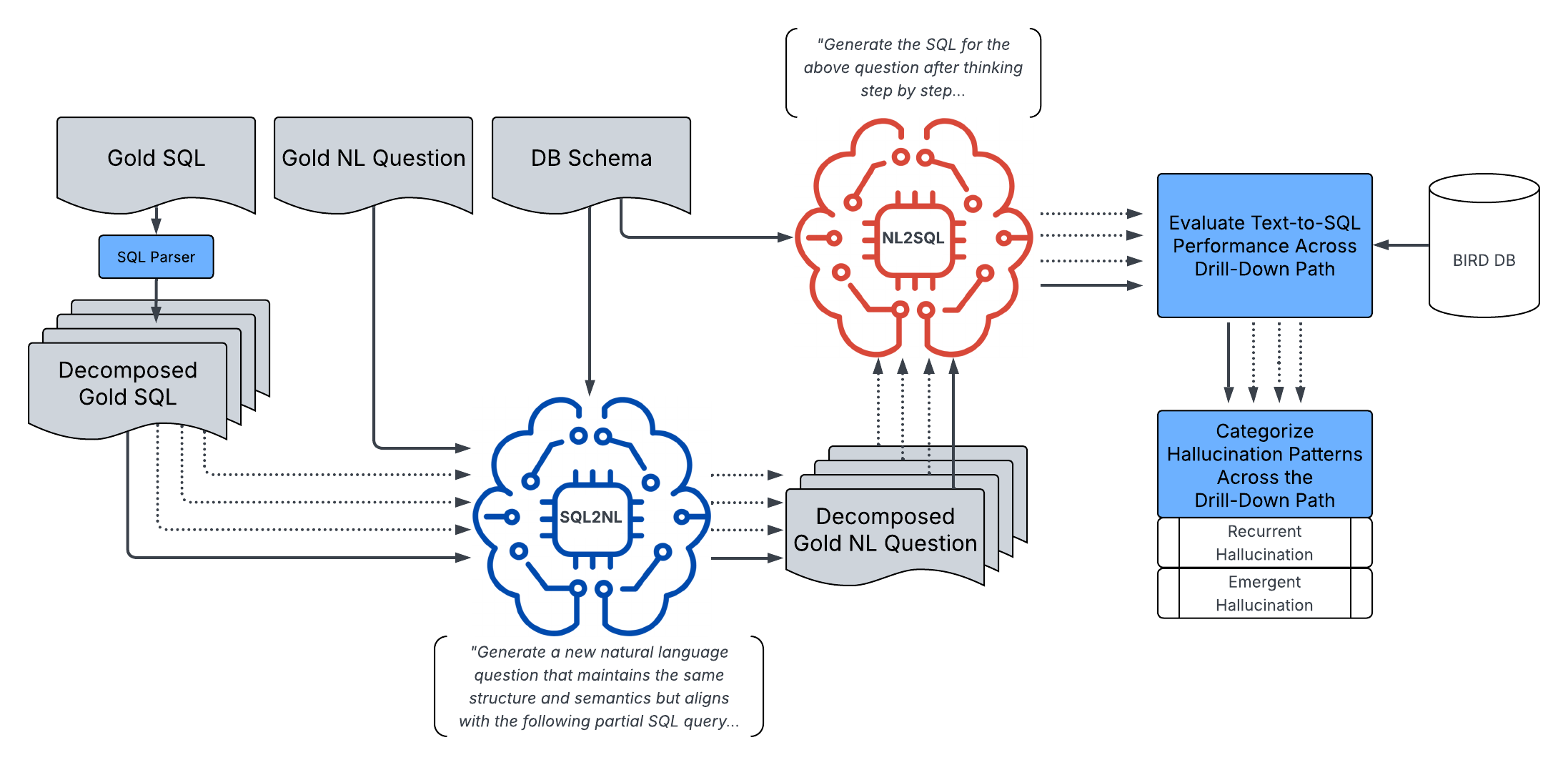}
\caption{Drill-down framework. Gold SQL is decomposed into progressively more complex executable sub-queries, which are paired with natural-language sub-questions. LLM outputs are evaluated at each step and annotated to trace recurrent and emergent hallucination patterns.}
\label{fig:drill-down}
\end{figure*}

We therefore study hallucinations descriptively across progressively more complex structured-generation steps. We decompose executable SQL queries from BIRD-mini into drill-down paths and pair each sub-query with a corresponding natural-language sub-question (Fig.~\ref{fig:drill-down}). We then evaluate five LLMs at each step and annotate hallucination categories using deterministic structural rules. Building on the hallucination taxonomy of \citet{qu2024before_generation_align_it}, we use a complementary temporal characterization: \textit{recurrent} hallucinations are observed earlier in the path and again in the final query, whereas \textit{emergent} hallucinations are observed only at the final step.

This framing is relevant to uncertainty-aware generation, but our experiments do not measure uncertainty itself. We do not estimate calibrated confidence, epistemic uncertainty, or predictive uncertainty, and we do not treat hallucination as a direct estimator or necessary consequence of uncertainty. Instead, we examine a reliability failure phenomenon that is relevant to uncertainty-aware systems and ask how its rule-based recurrence and emergence measures differ across progressive query steps and a conversational-history condition.

Our contributions are threefold:
\begin{enumerate}
    \item We present a drill-down evaluation framework for examining rule-based hallucination categories across progressively more complex Text-to-SQL generation steps.
    \item We use recurrent and emergent labels to distinguish categories observed both earlier and at the final step from categories observed only at the final step.
    \item Across five evaluated LLMs, we observe non-uniform temporal patterns: several categories show appreciable recurrence, clause abuse shows comparatively higher emergence, and the history condition is generally associated with higher recurrence and lower emergence.
\end{enumerate}

\section{Related Work}

Early Text-to-SQL systems commonly adopted sequence-to-sequence frameworks in which the natural-language question and target database schema were jointly encoded. Early work relied on recurrent architectures \citep{dong2016language, jia2016data}, followed by graph-based and relation-aware approaches \citep{bogin2019rat} and pre-trained transformer encoders \citep{yin2020tabert, yu2021grappa}. More recently, LLM-based approaches have become common in Text-to-SQL because of their few-shot learning capacity \citep{brown2020language}, in-context reasoning \citep{xie2021explanation}, and chain-of-thought prompting capabilities \citep{wei2022chain}. These approaches have achieved strong benchmark results while also raising hallucination and reliability concerns.

Hallucination research has developed taxonomies for unsupported or incorrect LLM outputs and has studied mitigation in Text-to-SQL specifically \citep{huang2023survey_hallucination, qu2024before_generation_align_it}. Our work is diagnostic rather than prescriptive: we do not introduce a new mitigation algorithm. Instead, we ask when hallucinations appear along a structured generation trajectory and whether they persist as the task becomes more complex.

Prior Text-to-SQL work has used decomposition largely as a generation strategy. CHASE-SQL \citep{pourreza2024chase_sql}, for example, applies divide-and-conquer reasoning to break complex problems into sub-components before synthesizing a final solution. Inspired by the decomposable structure of SQL, we use decomposition here as an analysis device across progressive sub-queries. This complements static hallucination categories with a descriptive view of whether the same category is observed earlier in a path and again at the final stage, or only at the final stage.

\section{Dataset}
\label{sec:Dataset}

\paragraph{BIRD-mini}
We conduct our experiments on the PostgreSQL \citep{postgresql} BIRD-mini dataset, a smaller version of the BIRD benchmark that reduces evaluation cost while retaining examples across multiple BIRD difficulty levels. Recent work on interactive Text-to-SQL and NL2SQL-agent settings provides related evaluation context \citep{interactiveT2S2025}. Applying our decomposition and annotation pipeline to the full BIRD development split would require many additional sub-questions and model generations across five commercial LLMs. We therefore use BIRD-mini rather than introducing an additional subsampling procedure.

Our decomposition expands the 500 original BIRD-mini examples into 1,384 executable question--query pairs. Each original query is decomposed as far as possible while preserving executable intermediate SQL. The resulting paths provide progressively more complex targets that can be evaluated independently and then related back to the corresponding final query.

The implementation can be applied to datasets beyond BIRD-mini, but the empirical frequencies reported here are specific to this evaluation. We do not assume that the prevalence or temporal distribution of hallucination categories would remain unchanged on the full BIRD benchmark or other Text-to-SQL datasets. We therefore interpret the results as descriptive patterns within the evaluated setting rather than as estimates of general or universal hallucination prevalence.

\section{Hallucination Characterization}

\paragraph{Schema-Based and Logic-Based}
We adopt the hallucination taxonomy of \citet{qu2024before_generation_align_it}, which separates errors into schema-based and logic-based categories. \textit{Schema-based hallucinations} concern the database structure or values, such as invalid schema references, unnecessary attributes, or value misrepresentation. \textit{Logic-based hallucinations} concern query construction, including unnecessary joins, clause abuse, or faulty mathematical/operator logic. Appendix~\ref{appendix:taxonomy}, Table~\ref{tab:hallucination-taxonomy}, provides the operational definitions used in our annotation pipeline.

\paragraph{Recurrent and Emergent Hallucinations}
Existing hallucination taxonomies describe the \textit{form} of an error. We use a complementary temporal characterization that describes how a rule-based category is observed across a drill-down path. A \textit{recurrent hallucination} is a category detected in at least one earlier step and again in the final query. An \textit{emergent hallucination} is a category detected in the final query with no earlier detection of that category in the same path. These labels are temporal abstractions over the underlying schema- and logic-based categories.

\paragraph{Hallucination Behavior Patterns}
We use recurrent and emergent labels as descriptive behavior patterns rather than as evidence of distinct causal mechanisms. A high recurrence probability means that, conditional on an instance of a category being observed earlier in the path, the same category is also observed in the final query relatively often. A high emergence probability means that the category is observed in the final query relatively often when no earlier instance of that category was detected. These conditional measures can motivate hypotheses about persistence or final-stage appearance, but they do not identify causal mechanisms or the fraction of all final-step errors that originated earlier.

\section{Methodology}

Our drill-down analysis consists of three stages:
\begin{enumerate}
    \item decompose BIRD-mini SQL into progressive executable sub-queries and pair them with natural-language sub-questions;
    \item evaluate multiple LLMs at each step of each drill-down path; and
    \item apply deterministic annotation rules to characterize hallucination type and temporal behavior (Fig.~\ref{fig:drill-down}).
\end{enumerate}

\subsection{Decompose and Generate Drill-Down Dataset}

\paragraph{Progressive Sub-Query Generation}
We decompose each BIRD-mini query into multiple progressive queries using an SQL parser \citep{sqlparse}. Parsing proceeds through the query structure so that each retained intermediate step remains executable. The resulting sequence gradually introduces additional conditions and structure until reaching the complete query.

\paragraph{Sub-Question Generation}
Each sub-query is paired with a natural-language sub-question describing what that executable SQL retrieves. All sub-queries are generated deterministically, whereas the natural-language realizations are generated with GPT-4o-mini using the database schema, original question, and target sub-query. We also regenerate the final natural-language question with the same procedure so that the questions within a path are produced under a consistent generation process.

This asymmetric design is motivated by prior work using formal-to-natural-language transformations to produce natural-language descriptions or semantically related formulations from structured inputs \citep{li2025evaluating, moore2025autoformalization}. This prior work motivates the design choice but does not establish that every generated sub-question in our dataset is semantically equivalent to its target SQL. The generated sub-questions were not independently validated against manually authored equivalents. Consequently, some observed Text-to-SQL behavior may reflect artifacts introduced by the SQL-to-NL transformation. We treat this as a potential source of variation in the evaluation and discuss it explicitly in the limitations.

\subsection{Drill-Down Evaluation and Hallucination Annotation}
Following decomposition, each model generates SQL for every natural-language question in the progressive path. We compare the predicted SQL against the corresponding ground-truth SQL and database schema. The deterministic annotation pipeline records six hallucination categories and then determines whether each final-step hallucination is recurrent or emergent. Appendix~\ref{appendix:annotation-rules} specifies the exact structural rules.

These labels should be interpreted as \textit{operational indicators} rather than human-validated semantic gold labels. Deterministic rules provide consistent application across models and steps, but they cannot capture every semantic equivalence or error mode that a human SQL expert might identify.

\subsection{Reproducibility Artifacts}
\label{sec:repro}
We release the decomposition code, model-prompting code, generated outputs, hallucination annotations, evaluation code, and figure-generation scripts in the repository linked in the abstract. The repository is intended to make the processing pipeline and pre-generated results inspectable without requiring reviewers to rerun commercial-model API calls. The manuscript remains self-contained; the repository provides implementation and replication artifacts rather than evidence required to understand the claims.

\section{Experimental Setup and Evaluation}
\label{subsec:experiment}

We evaluate five LLMs: Claude-3.5-Sonnet and Claude-3.7-Sonnet from Anthropic \citep{anthropic2024claude35, anthropic2025claude37}, and GPT-4-Turbo, GPT-4o-mini, and GPT-4.1-mini from OpenAI \citep{openai2023gpt4, openai2024gpt4o, openai2025gpt41mini}. Models are evaluated on the BIRD-mini drill-down dataset using the BIRD Text-to-SQL prompt shown in Appendix~\ref{appendix:prompts}. For every step, we compare predicted SQL, ground-truth SQL, and the corresponding database schema.

The experiments address three research questions:

\paragraph{RQ1: Hallucination persistence}
When hallucinations occur in simpler stages of a decomposed Text-to-SQL problem, how frequently do they reappear in the final query?

\paragraph{RQ2: Hallucination emergence}
Which hallucination categories show comparatively higher conditional emergence at the final stage when no earlier instance of the same category is detected?

\paragraph{RQ3: Contextual history}
How do recurrence and emergence probabilities differ when models do or do not have access to earlier questions and generations in a decomposed path?

\subsection{Evaluation Metrics}

We adopt the problem formulation from \citet{qu2024before_generation_align_it}. Given a natural-language question $\mathcal{Q}$ and associated database schema $\mathcal{D} = \langle \mathcal{C}, \mathcal{T} \rangle$, where $\mathcal{C}$ and $\mathcal{T}$ represent the column and table sets, respectively, the Text-to-SQL task is to generate a valid SQL query $y$ that reflects the intent encoded in $\mathcal{Q}$.

\paragraph{Execution Accuracy (EX)}
We evaluate baseline model performance using \textit{Execution Accuracy (EX)} \citep{li2023bird}, which measures whether a predicted SQL query $\hat{y}$ yields the same execution result as ground truth query $y^*$ on database $\mathcal{D}$:
\[
\text{EX}(\hat{y}, y^*) =
\begin{cases}
1 & \text{if } \texttt{Exec}(\hat{y}, \mathcal{D}) = \texttt{Exec}(y^*, \mathcal{D}) \\
0 & \text{otherwise.}
\end{cases}
\]
Across $N$ examples,
\[
\text{EX}_{\text{avg}} = \frac{1}{N} \sum_{i=1}^{N} \text{EX}(\hat{y}^{(i)}, y^{*(i)}).
\]

\paragraph{Soft-F1 Score}
We also report the \textit{Soft-F1 Score} \citep{li2023bird}, which gives partial credit when the predicted and ground-truth execution results overlap. Let $\hat{T} = \texttt{Exec}(\hat{y}, \mathcal{D})$ and $T^* = \texttt{Exec}(y^*, \mathcal{D})$. With tuple-level true positives (TP), false positives (FP), and false negatives (FN),
\[
\text{Soft-F1} = \frac{2 \cdot \text{TP}}{2 \cdot \text{TP} + \text{FP} + \text{FN}}.
\]

\section{Hallucination Dynamics}
\label{sec:results}

\begin{figure}[t]
\centering
\includegraphics[width=\linewidth]{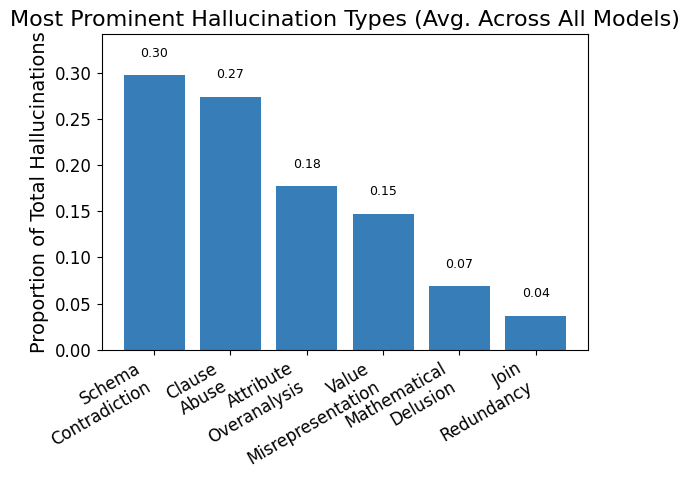}
\caption{Distribution of rule-based hallucination categories across the evaluated outputs, averaged across models. Schema contradiction and clause abuse are the most common categories in this analysis.}
\label{fig:distribution}
\end{figure}

\begin{figure*}[t]
\centering
\begin{subfigure}[t]{0.48\linewidth}
    \centering
    \includegraphics[width=\linewidth]{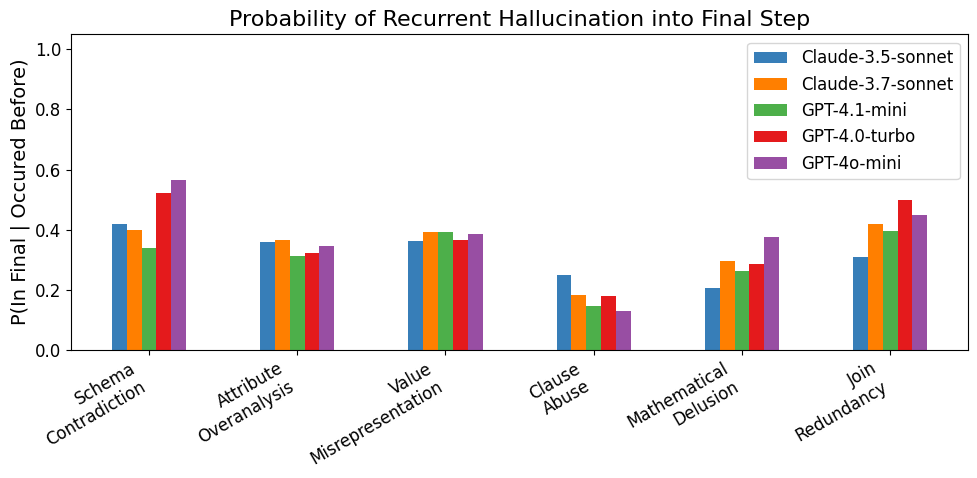}
    \caption{Recurrent hallucinations}
    \label{fig:bar-side-a}
\end{subfigure}
\hfill
\begin{subfigure}[t]{0.48\linewidth}
    \centering
    \includegraphics[width=\linewidth]{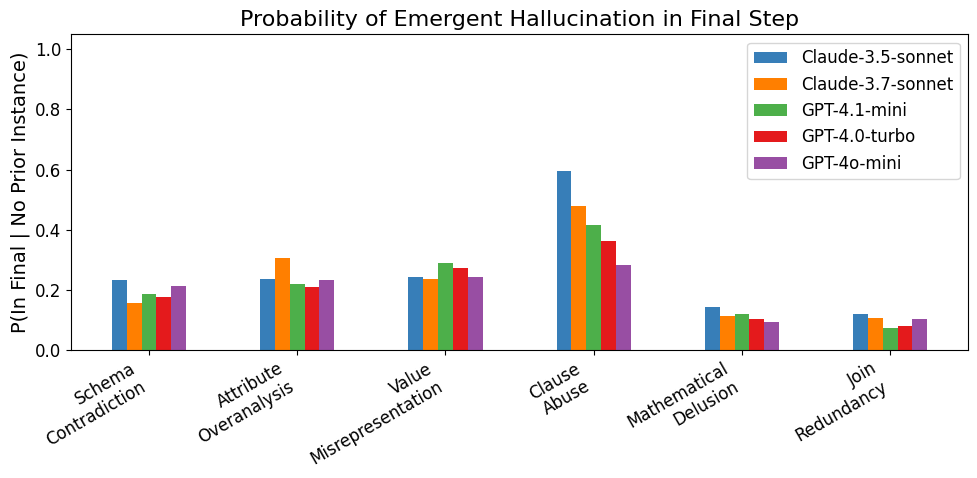}
    \caption{Emergent hallucinations}
    \label{fig:bar-side-b}
\end{subfigure}
\caption{Conditional probabilities for recurrent (a) and emergent (b) rule-based hallucination categories. For most categories, $P(\text{In Final Step}\mid\text{Occurs Earlier})$ is higher than $P(\text{In Final Step}\mid\text{Does Not Occur Earlier})$; clause abuse shows a different pattern.}
\label{fig:bar-side}
\end{figure*}

To validate the overall Text-to-SQL setup, we first report BIRD-mini execution performance in Appendix~\ref{appendix:exp-results}, Table~\ref{tab:bird-mini}. Figure~\ref{fig:distribution} summarizes the hallucination categories observed across the evaluated outputs.

\subsection{Recurrent Hallucinations}

Figure~\ref{fig:bar-side-a} reports $P(\text{In Final Step}\mid\text{Occurs in Earlier Steps})$: the probability that a hallucination category is present in the final query given that the same category was detected earlier in the drill-down path. Several categories show appreciable recurrence probabilities, indicating that an earlier instance of a category can be followed by an instance of the same category in the final query. Schema contradiction is common overall and has relatively high recurrence probabilities in the evaluated models. Clause abuse has lower recurrence than several other categories.

These results answer RQ1 descriptively: for several categories, an error detected earlier in a path has a nontrivial conditional probability of being detected again in the final query. Importantly, $P(\text{In Final Step}\mid\text{Occurs Earlier})$ does not measure the fraction of all final-step errors that were already present earlier, and the analysis does not establish that an earlier hallucination \textit{causes} a later one.

\subsection{Emergent Hallucinations}

Figure~\ref{fig:bar-side-b} reports $P(\text{In Final Step}\mid\text{Does Not Occur Earlier})$: the probability that a category appears in the final query when no instance of that same category was detected earlier in the path. These emergence probabilities are lower than the corresponding recurrence probabilities for most categories. Clause abuse differs from this pattern and shows comparatively high emergence probabilities.

For RQ2, clause abuse differs from the pattern observed for several other categories. One possible explanation is that the final query contains more opportunities for additional global clauses or logical operators. This interpretation is speculative: the present experiment does not isolate which property of the final step, if any, is responsible for the observed pattern.

\begin{figure*}[t]
\centering
\includegraphics[width=0.95\textwidth]{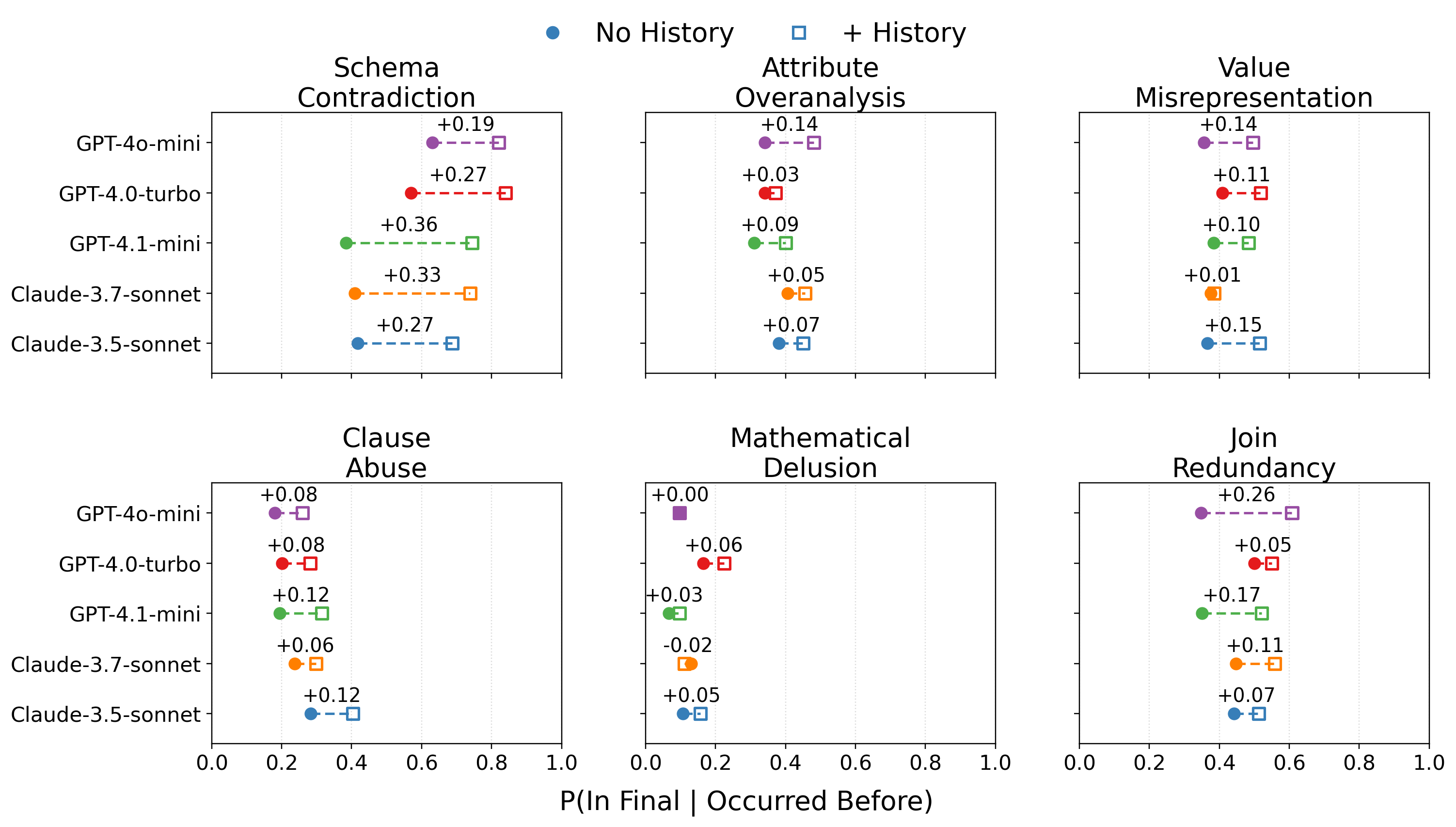}
\caption{Probability of recurrent hallucinations with and without conversational history. Across most evaluated model--category pairs, the history condition is associated with higher recurrence probabilities, with some differences close to zero.}
\label{fig:recurrent-2}
\end{figure*}

\begin{figure*}[t]
\centering
\includegraphics[width=0.95\textwidth]{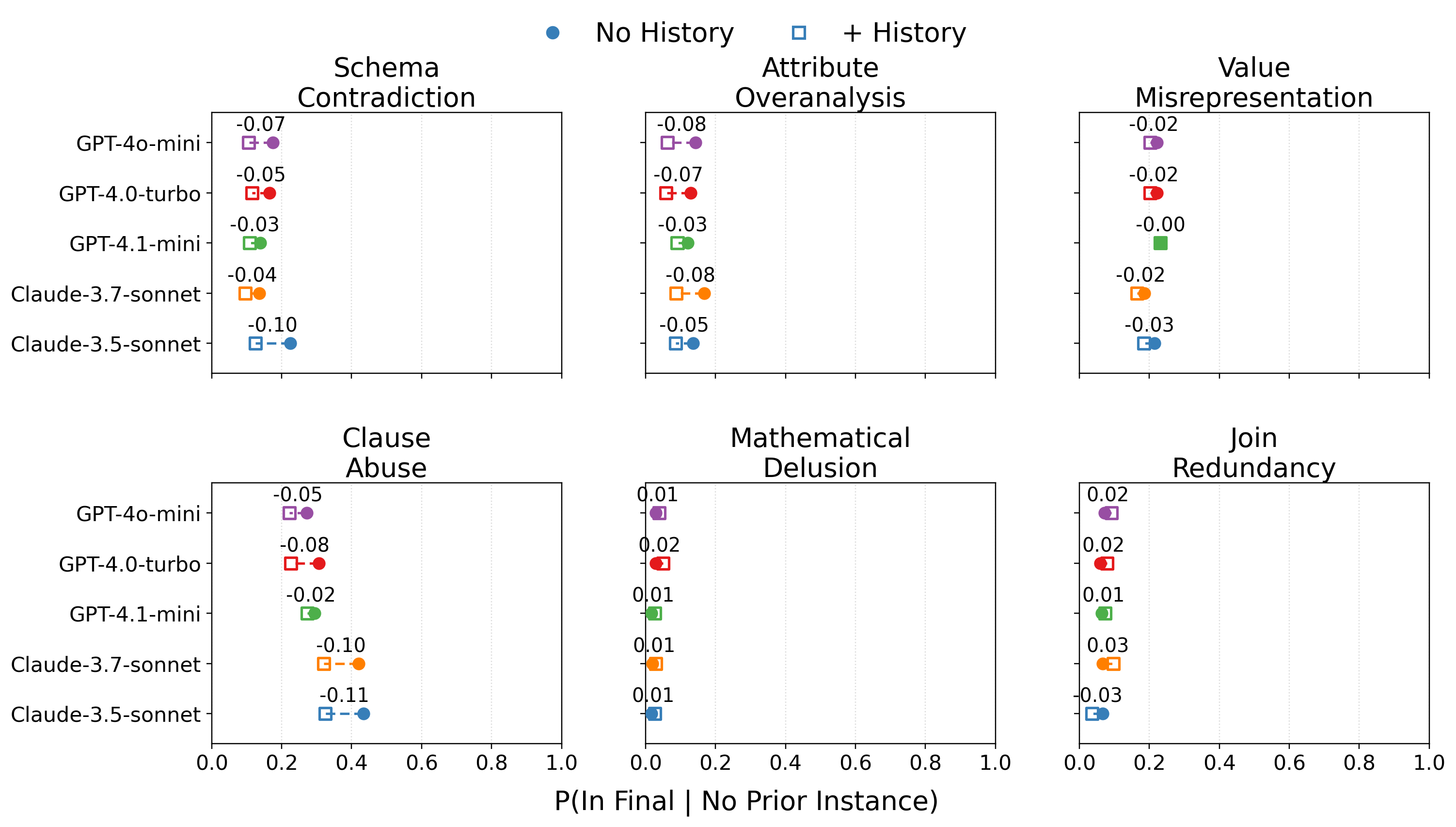}
\caption{Probability of emergent hallucinations with and without conversational history. Across most evaluated model--category pairs, the history condition is associated with lower emergence probabilities, with some differences close to zero.}
\label{fig:emergent-2}
\end{figure*}

\subsection{Contextual-History Trade-off}

Figures~\ref{fig:recurrent-2}--\ref{fig:emergent-2} compare generation with and without access to prior conversational history from the drill-down path. The two measures generally shift in opposite directions. Under the history condition, recurrent hallucination probabilities are higher in most evaluated model--category pairs, while emergent hallucination probabilities are lower in most pairs. The magnitude of these differences varies considerably across models and categories, and some differences are close to zero.

For RQ3, the results do not support a simple claim that the history condition is uniformly better or worse. Rather, the history condition is associated with a different pattern across the two measures: emergence probabilities are generally lower, whereas recurrence probabilities are generally higher. This suggests that adding prior context should not be assumed to reduce every measured hallucination behavior.

\subsection{Task Performance}

Among the evaluated models, GPT-4.1-mini has the highest total EX (42.8\%) and Soft-F1 (46.53\%), while GPT-4o-mini has the lowest totals (32.8\% EX and 36.48\% Soft-F1). Performance varies by BIRD difficulty level. For GPT-4.1-mini without history, for example, EX is 59.46\% on simple questions and 21.57\% on challenging questions. Averaging the model-level totals in Table~\ref{tab:bird-mini}, mean EX is 38.28\% without history and 37.56\% with history, a difference of $-0.72$ percentage points; mean Soft-F1 is 42.50\% without history and 41.41\% with history, a difference of approximately $-1.09$ points. The direction varies by model: Claude-3.7-Sonnet increases from 38.8 to 39.8 EX with history, whereas GPT-4.1-mini decreases from 42.8 to 40.0 EX. Full results are provided in Appendix~\ref{appendix:exp-results}.

\section{Interpreting Hallucination Dynamics}
\label{sec:discussion}

The drill-down analysis provides a trajectory-level description in addition to final-output evaluation. For RQ1, the recurrence probabilities show that categories detected earlier can also be detected again in the final query with appreciable probability for several error types. This pattern is compatible with multiple explanations, including persistent misinterpretation, but the present analysis cannot distinguish among them. For RQ2, clause abuse has comparatively high emergence probabilities, which may reflect properties of final-query composition, although that explanation is not directly tested. For RQ3, the history condition is generally associated with higher recurrence and lower emergence probabilities, suggesting that conclusions about context depend on which hallucination measure is considered.

\subsection{Relationship to Uncertainty-Aware Generation}

Our study does not measure calibrated confidence, predictive entropy, or epistemic uncertainty, and we therefore do not interpret the reported probabilities as model uncertainty estimates or claim that the detected hallucinations are caused by uncertainty. Instead, we characterize a reliability failure phenomenon that is relevant to uncertainty-aware generation. The analysis varies two aspects of the evaluation setting: position within a progressively constructed query path and access to prior conversational history. Under the history condition, recurrence probabilities are generally higher and emergence probabilities are generally lower, so the amount of available context alone should not be treated as a proxy for lower values on both hallucination measures.

The generated sub-questions introduce an additional potential source of variation in the experimental pipeline. Because these natural-language realizations are generated rather than manually authored, variation in wording may alter downstream model behavior. The SQL targets remain deterministic, but the observed hallucination patterns therefore characterize the complete SQL-to-NL-to-SQL evaluation pipeline rather than a controlled set of human-authored paraphrases.

\subsection{Design Implications}

The observed patterns suggest several hypotheses that could be tested in future mitigation work, but none is evaluated as an intervention here. The recurrence of schema contradiction suggests examining whether early schema checks could reduce later errors. The comparatively high emergence of clause abuse suggests testing final-query structural verification. The history results also suggest testing selective or verified history rather than assuming that retaining all prior generations is beneficial. These are design hypotheses only and require controlled experiments before any effectiveness claim can be made.

\section{Conclusion}

We present a drill-down evaluation of recurrent and emergent hallucination patterns in Text-to-SQL generation. Rather than evaluating only the final query, we examine rule-based hallucination categories across progressively more complex executable sub-queries. Across five LLMs, several categories show appreciable recurrence probabilities, meaning that a category detected earlier can also be detected again in the final query. Clause abuse differs from several other categories by showing comparatively high emergence probabilities.

We also observe that the conversational-history condition is generally associated with lower emergence probabilities and higher recurrence probabilities, although the magnitude varies by model and category. These results suggest value in evaluating hallucination behavior across generation trajectories rather than relying only on final-output metrics, and they motivate future controlled work on verification, context selection, and hallucination mitigation in Text-to-SQL systems.

\section*{Limitations}

Our analysis has several limitations. First, all reported experiments use BIRD-mini. The decomposition expands the benchmark into 1,384 executable question--query pairs, but the prevalence and temporal distribution of hallucination categories should not be assumed to generalize unchanged to the full BIRD benchmark, Spider, other database domains, or other structured-generation tasks.

Second, we evaluate five closed-source commercial LLMs. We release the generated outputs and evaluation artifacts to support inspection and reuse, but exact regeneration depends on external APIs and model versions that may change over time.

Third, the natural-language sub-questions are generated by GPT-4o-mini and were not independently validated against manually authored equivalents. Although the executable SQL targets are generated deterministically and prior work provides some support for using the SQL-to-NL direction, some measured behavior may reflect artifacts of the generated natural-language realizations. A controlled comparison with manually authored sub-questions would provide a stronger validity check but would require additional data collection beyond the present study.

Fourth, hallucination categories are assigned with deterministic rule-based heuristics rather than large-scale human annotation. The rules provide consistent operational labels across models and steps, but they may miss semantic equivalences, misclassify unusual SQL constructions, or fail to capture error types outside the predefined taxonomy. The resulting labels should therefore be interpreted as rule-based indicators rather than definitive human judgments.

Finally, recurrent and emergent hallucinations are descriptive temporal categories based on conditional probabilities. The experiments document associations across generation stages and history conditions but do not identify causal mechanisms. In particular, the recurrence metric should not be interpreted as the proportion of final-step errors that originated earlier. The mitigation implications discussed above are hypotheses motivated by the observed patterns, not experimentally verified defenses.


\bibliography{custom}

\appendix

\section{Taxonomy of Hallucination Types}
\label{appendix:taxonomy}

\begin{table}[H]
\centering
\small
\begin{tabular}{p{0.30\linewidth} p{0.62\linewidth}}
\toprule
\textbf{Category} & \textbf{Description} \\
\midrule
Schema-Based: Schema Contradiction &
The rule flags predicted tables, columns, or aliases not supported by the database schema, together with selected wildcard or backtick patterns handled by the implementation. \\
\midrule
Schema-Based: Attribute Overanalysis &
The rule flags additional valid tables or columns that appear in the prediction but not in the reference SQL, serving as an operational indicator of possible over-specification. \\
\midrule
Schema-Based: Value Misrepresentation &
The rule flags selected differences in literal values or type-casting behavior between the prediction and reference SQL. \\
\midrule
Logic-Based: Join Redundancy &
The rule flags predictions containing more JOIN operations than the reference SQL, indicating additional relational structure under this operational definition. \\
\midrule
Logic-Based: Clause Abuse &
The rule flags selected structural SQL clauses (e.g., \texttt{GROUP BY}, \texttt{LIMIT}, \texttt{ORDER BY}) or logical operators (e.g., \texttt{AND}, \texttt{OR}) that appear in the prediction but not in the reference SQL. \\
\midrule
Logic-Based: Mathematical Delusion &
The rule flags selected mathematical or operator patterns, including division without casting, use of \texttt{\%}, selected uses of \texttt{BETWEEN}, or parser-detected malformed arithmetic structures. \\
\bottomrule
\end{tabular}
\caption{Operational taxonomy of hallucination types, adapted from \citet{qu2024before_generation_align_it}.}
\label{tab:hallucination-taxonomy}
\end{table}

\begin{table}[H]
\centering
\small
\begin{tabular}{p{0.30\linewidth} p{0.62\linewidth}}
\toprule
\textbf{Category} & \textbf{Description} \\
\midrule
Recurrent Hallucination &
A rule-based hallucination category detected in at least one earlier step and again in the final step. \\
\midrule
Emergent Hallucination &
A rule-based hallucination category detected in the final step with no earlier detection of that category in the same path. \\
\bottomrule
\end{tabular}
\caption{Recurrent and emergent hallucination definitions used in the drill-down analysis.}
\label{tab:rec-emergent}
\end{table}

\section{Hallucination Annotation Rules}
\label{appendix:annotation-rules}

This appendix details the deterministic rule-based procedure used to annotate hallucination categories in predicted SQL queries. Each predicted query is compared against its ground-truth SQL and the corresponding BIRD schema. These rules correspond to the implementation used during evaluation.

\subsection{Annotation Pipeline Overview}

For each step in a drill-down path, we execute the following procedure:
\begin{enumerate}
    \item \textbf{Parse SQL structure:} Extract tables, columns, aliases, alias--column pairs, SELECT elements, JOIN structures, and literal values from both the ground-truth and predicted queries.
    \item \textbf{Normalize queries:} Convert SQL strings to normalized lowercase/uppercase variants as required by rule-specific pattern matching.
    \item \textbf{Compare predicted SQL to the schema and ground truth:} Validate tables, columns, aliases, clauses, and operators against the database schema, the ground-truth SQL query, and SQL structural constraints.
    \item \textbf{Assign hallucination categories:} Map detected violations to one of six main hallucination types.
    \item \textbf{Deduplicate:} Count each main hallucination type at most once per query step, while retaining fine-grained subcategory descriptions.
\end{enumerate}

We use the taxonomy described in the main paper: Schema-Based (Schema Contradiction, Attribute Overanalysis, Value Misrepresentation) and Logic-Based (Clause Abuse, Mathematical Delusion, Join Redundancy).

\subsection{Schema-Based Hallucinations}

\subsubsection{Schema Contradiction}
A prediction is labeled as a \textbf{Schema Contradiction} when the implementation detects one or more of the following schema-related patterns:
\begin{itemize}
    \item \textbf{Unknown tables:} use of tables not present in the database schema.
    \item \textbf{Unknown columns:} use of columns that do not appear in any table referenced in the query.
    \item \textbf{Alias errors:} an alias refers to a nonexistent table, or an aliased column does not exist in the referenced table.
    \item \textbf{Wildcard or backtick pattern:} presence of patterns such as \texttt{table.*} or MySQL-style backticks that are flagged by the implementation.
    \item \textbf{Missing table reference:} a column exists in the schema but is used without including its table in the FROM or JOIN clauses.
\end{itemize}

\subsubsection{Attribute Overanalysis}
A prediction is labeled as \textbf{Attribute Overanalysis} when it contains additional valid schema elements that do not appear in the reference SQL:
\begin{itemize}
    \item extra columns not used in the ground-truth query;
    \item extra tables joined relative to the reference SQL.
\end{itemize}
This category is used as an operational indicator of possible over-specification rather than invalid schema references.

\subsubsection{Value Misrepresentation}
A prediction is labeled as \textbf{Value Misrepresentation} when the rule detects selected differences in literal values or type-casting behavior, such as:
\begin{itemize}
    \item mismatched or altered literal values in the WHERE clause;
    \item casts present in the prediction but absent in the reference SQL;
    \item missing casts that appear in the ground truth.
\end{itemize}

\subsection{Logic-Based Hallucinations}

\subsubsection{Clause Abuse}
A prediction is labeled as \textbf{Clause Abuse} when the rule detects selected structural SQL clauses or logical constructs in the prediction that do not appear in the reference SQL, including:
\begin{itemize}
    \item extra top-level clauses such as \texttt{GROUP BY}, \texttt{HAVING}, \texttt{ORDER BY}, \texttt{LIMIT}, or \texttt{OFFSET};
    \item additional logical operators such as \texttt{OR} or additional uses of \texttt{AND};
    \item introduction of JOIN variants, CTEs, set operations, or vendor-specific syntax not present in the ground truth.
\end{itemize}

\subsubsection{Mathematical Delusion}
A prediction is labeled as \textbf{Mathematical Delusion} when the implementation detects selected mathematical or operator patterns, such as:
\begin{itemize}
    \item integer division without casting;
    \item use of the modulo operator (\texttt{\%}) in patterns treated by the rule as percentage-like;
    \item selected uses of \texttt{BETWEEN} outside the numeric/date patterns expected by the rule;
    \item malformed or incomplete operator structures identified through SQL parser errors.
\end{itemize}

\subsubsection{Join Redundancy}
A prediction is labeled as \textbf{Join Redundancy} when it contains more JOIN operations than the reference SQL, indicating additional relational structure under this operational definition.

\subsection{Deduplication Procedure}
For each query step:
\begin{itemize}
    \item each main hallucination type is counted at most once;
    \item all subcategory descriptions are recorded for fine-grained analysis; and
    \item an annotated list of hallucinations is stored in the evaluation JSON.
\end{itemize}

The deterministic procedure provides consistent operational labels across models and drill-down steps. Because these heuristics have not been comprehensively validated against human judgments, we treat the resulting categories as rule-based indicators of hallucination patterns rather than definitive semantic annotations.

\section{Prompts}
\label{appendix:prompts}

\paragraph{Prompt 1: Progressive Question Rewriting Prompt}
\begin{Verbatim}[breaklines=true,breakanywhere=true,fontsize=\small]
Let's take this step-by-step.

Given this database schema: 
{schema_prompt}

Given this original question: 
"{original_question}"

Generate a new natural language question 
that maintains the same structure and 
semantics but aligns with the following 
SQL query:
{partial_sql}

Do not include any information in your 
generated question that is not directly 
included in the query. The original 
question should be used as reference to 
generate this question.

Requirements:
- The generated question must correspond 
exactly to what this SQL retrieves
- Maintain the same domain context and 
terminology as the original question
- The question should be answerable using 
only this SQL query

Generate only the natural language question.
\end{Verbatim}

\paragraph{Prompt 2: Text-to-SQL Prompt}
\begin{Verbatim}[breaklines=true,breakanywhere=true,fontsize=\small]
Using valid {sql_dialect} and 
understanding External Knowledge:
{knowledge}

{base_prompt}{knowledge_text}, answer 
the following questions for the tables 
provided above. Generate the {sql_dialect} 
for the above question after thinking 
step by step:

In your response, you do not need to 
mention your intermediate steps. Do not 
include any comments in your response.
Do not need to start with the symbol ``
You only need to return the result 
{sql_dialect} SQL code start from SELECT
\end{Verbatim}

\section{Experiment Results}
\label{appendix:exp-results}

\begin{table}[H]
\centering
\scriptsize
\setlength{\tabcolsep}{2.5pt}
\renewcommand{\arraystretch}{0.95}

\resizebox{\columnwidth}{!}{%
\begin{tabular}{@{}lcccc@{}}
\toprule
\textbf{Model} & \textbf{Simple} & \textbf{Moderate} &
\textbf{Challenging} & \textbf{Total} \\
\midrule
\textit{Count} & 148 & 250 & 102 & 500 \\
\midrule

Claude-3.5-Sonnet
& 56.08 & 35.20 & 23.53 & 39.00 (EX) \\
& 59.39 & 38.61 & 31.15 & 43.24 (F1) \\

Claude-3.5-Sonnet (+ History)
& 50.00 & 34.40 & 20.59 & 36.20 (EX) \\
& 56.93 & 37.84 & 29.01 & 41.69 (F1) \\
\midrule

Claude-3.7-Sonnet
& 51.35 & 38.40 & 21.57 & 38.80 (EX) \\
& 55.09 & 43.58 & 29.59 & 44.13 (F1) \\

Claude-3.7-Sonnet (+ History)
& 52.70 & 38.40 & 24.51 & 39.80 (EX) \\
& 57.32 & 42.10 & 30.78 & 44.30 (F1) \\
\midrule

GPT-4-Turbo
& 58.78 & 34.00 & 17.65 & 38.00 (EX) \\
& 60.66 & 38.45 & 24.14 & 42.11 (F1) \\

GPT-4-Turbo (+ History)
& 55.41 & 37.60 & 19.61 & 39.20 (EX) \\
& 57.45 & 40.48 & 25.16 & 42.38 (F1) \\
\midrule

GPT-4o-mini
& 47.97 & 31.60 & 13.73 & 32.80 (EX) \\
& 50.87 & 34.42 & 20.63 & 36.48 (F1) \\

GPT-4o-mini (+ History)
& 47.30 & 31.20 & 14.71 & 32.60 (EX) \\
& 48.82 & 33.57 & 20.10 & 35.34 (F1) \\
\midrule

GPT-4.1-mini
& 59.46 & 41.60 & 21.57 & 42.80 (EX) \\
& 61.47 & 45.04 & 28.48 & 46.53 (F1) \\

GPT-4.1-mini (+ History)
& 56.08 & 39.20 & 18.63 & 40.00 (EX) \\
& 58.33 & 42.33 & 24.09 & 43.35 (F1) \\

\bottomrule
\end{tabular}%
}

\caption{BIRD-mini execution accuracy (EX, \%) and Soft-F1 scores across difficulty levels.}
\label{tab:bird-mini}
\end{table}

\end{document}